\documentclass[runningheads]{llncs}
\usepackage{graphicx}
\usepackage{tikz}
\usepackage{comment}
\usepackage{amsmath,amssymb} % define this before the line numbering.
\usepackage{color}
\usepackage{amssymb}% http://ctan.org/pkg/amssymb
\usepackage{pifont}% ht

\usepackage{color}
\usepackage{epsfig}
\usepackage{graphicx}
\usepackage{multirow}
\usepackage{graphics}
\usepackage{amsmath}
\usepackage{amssymb}
\usepackage{multirow}
\usepackage{algorithm}
\usepackage{algorithmicx}
\usepackage{subcaption}
\usepackage{sidecap}
\usepackage{wrapfig}

\usepackage{colortbl}
\usepackage{arydshln,graphicx,xcolor,array}
\usepackage{multirow}
\newcommand{\cmark}{\ding{51}}%
\newcommand{\xmark}{\ding{55}}%
\newcommand{\squeezeup}{\vspace{-1.5cm}}
\newcommand{\smallsqueezeup}{\vspace{-0.8cm}}
\newcommand{\tinysqueezeup}{\vspace{-0.4cm}}

\usepackage[utf8]{inputenc}
\let\origtau\tau % sav
\renewcommand{\tau}{\scalebox{1.2}{$\origtau$}}
\usepackage[colorlinks=true,linkcolor=red]{hyperref}%

\usepackage[accsupp]{axessibility}  % Improves PDF readability for those with disabilities.

\begin{document}
% \linenumbers
\pagestyle{headings}
\mainmatter
\def\ECCVSubNumber{852}  % Insert your submission number here

\title{COUCH: Towards Controllable Human-Chair Interactions} % Replace with your title

\begin{comment}
% \titlerunning{ECCV-22 submission ID \ECCVSubNumber} 
% \authorrunning{ECCV-22 submission ID \ECCVSubNumber} 
% \author{Anonymous ECCV submission}
% \institute{Paper ID \ECCVSubNumber}
\end{comment}
%******************

% \begin{comment}
\titlerunning{COUCH: Towards Controllable Human-Chair Interactions}
\author{Xiaohan Zhang\inst{1,2} \and
Bharat Lal Bhatnagar\inst{1,2} \and
Vladimir Guzov \inst{1,2}\and \\
Sebastian Starke\inst{3,4} \and
Gerard Pons-Moll\inst{1,2}}
\authorrunning{Zhang et al.}
\institute{University of Tübingen, Germany\\
% \email{\{abc,lncs\}@uni-tuebingen.de}
\and
Max Planck Institute for Informatics, Saarland Informatics Campus, Germany \\
\and 
Electronic Arts \\
\and 
University of Edinburgh, United Kingdom}

% \email{\{abc,lncs\}@mpi-inf.mpg.de}}
% \and
% Max Planck Institute for Informatics, Saarland Informatics Campus, Germany\\
% \email{\{abc,lncs\}@mpi-inf.mpg.de}}

% \end{comment}
%******************
\maketitle

%% Type definitions
% \usepackage{amsmath}
% \DeclareMathOperator*{\argmax}{arg\,max}

\newcommand{\model}[0]{COUCH}
\renewcommand{\vec}[1]{\boldsymbol{#1}}
\newcommand{\mat}[1]{\mathbf{#1}}
\newcommand{\set}[1]{\mathcal{#1}}
\newcommand{\vect}[1]{\mathbf{#1}}

%% SMPL definitions
\newcommand{\template}[0]{\mat{T}}
\newcommand{\blendweight}[0]{w}
\newcommand{\blendweights}[0]{\mat{W}}

\newcommand{\shape}[0]{\vec{\beta}}
\newcommand{\trans}[0]{\vect{t}}
\newcommand{\joints}[0]{\mat{J}}
\newcommand{\offsets}[0]{\mathbf{D}}

\newcommand{\smpl}[0]{M}
\newcommand{\posefun}[0]{T}
\newcommand{\blendfun}[0]{W}
\newcommand{\offsetfun}[0]{B}
\newcommand{\jointfun}[0]{J}

\newcommand{\hist}[1]{{#1}^-_t}
\newcommand{\future}[1]{{#1}^+_t}

%% Commenting definitions
\newcommand{\BB}[1]{{\textcolor{purple}{[\textbf{BB:} #1]}}}
\newcommand{\GPM}[1]{{\textcolor{red}{[\textbf{GPM:} #1]}}}
\newcommand{\XZ}[1]{{\textcolor{blue}{[\textbf{XZ:} #1]}}}
\newcommand{\Seb}[1]{{\textcolor{magenta}{[\textbf{SS:} #1]}}}

%% Project specific definitions
%% Hand Control
\newcommand{\control}{\mat{C}} 
\newcommand{\contact}{\vec{c}}
\newcommand{\handtraj}{\mat{h}}
\newcommand{\handtrajinterp}{\mat{\tilde{h}}}
\newcommand{\localphase}{\mat{\phi}}
\newcommand{\globalphase}{\mat{\Phi}}

\newcommand{\handtrajpred}{{\dot{\mat{h}}}}
\newcommand{\localphasepred}{\dot{\mat{\phi}}}
\newcommand{\globalphasepred}{\dot{\mat{\Phi}}}

%% Motion Network
\newcommand{\poseposition}{\vec{j}^{p}_i}
\newcommand{\posevelocity}{\vec{j}^{v}_{i}}
\newcommand{\poserotation}{\vec{j}^{r}_{i}}
\newcommand{\poseacc}{\ddot{\vec{j}}_{i}}
\newcommand{\feettar}{\tilde{\vec{f}}_{i}}

\newcommand{\trajpos}{\vec{t}^{p}_{i}}
\newcommand{\trajdir}{\vec{t}^{d}_{i}}
\newcommand{\trajact}{\vec{t}^{a}_{i}}

\newcommand{\goalpos}{\vec{g}^{p}_{i}}
\newcommand{\goaldir}{\vec{g}^{d}_{i}}
\newcommand{\goalact}{\vec{g}^{a}_{i}}

\newcommand{\trajposgoal}{\tilde{\vec{t}}^{p}_{i}}
\newcommand{\trajdirgoal}{\tilde{\vec{t}}^{d}_{i}}

\newcommand{\pose}{\mat{J}}
\newcommand{\traj}{\mat{T}}
\newcommand{\goal}{\mat{G}}
\newcommand{\environment}{\mat{E}}
\newcommand{\scene}{\mat{I}}
\newcommand{\phase}{\phi}

% additional
\newcommand{\futureposeposition}{\vec{\tilde{j}}^{p}_{i+1}}
\newcommand{\goaltrajdir}{\vec{\tilde{t}}^{d}_{i+1}}
\newcommand{\goaltrajpos}{\vec{\tilde{t}}^{p}_{i+1}}

\newcommand{\goaltraj}{\mat{\tilde{T}}}
\newcommand{\binarycontact}{\vec{b}_{i+1}}

\newcommand{\x}{\mat{X}}
\newcommand{\y}{\mat{Y}}

\newcommand{\latent}{\vec{z}}

\newcommand{\expertweights}{\mat{\alpha}}

\newcommand{\myparagraph}[1]{\vspace{1pt}\noindent{\bf #1}}

\smallsqueezeup

% teaser figure
\begin{figure}[h]
\begin{center}
\includegraphics[width=0.8\linewidth]{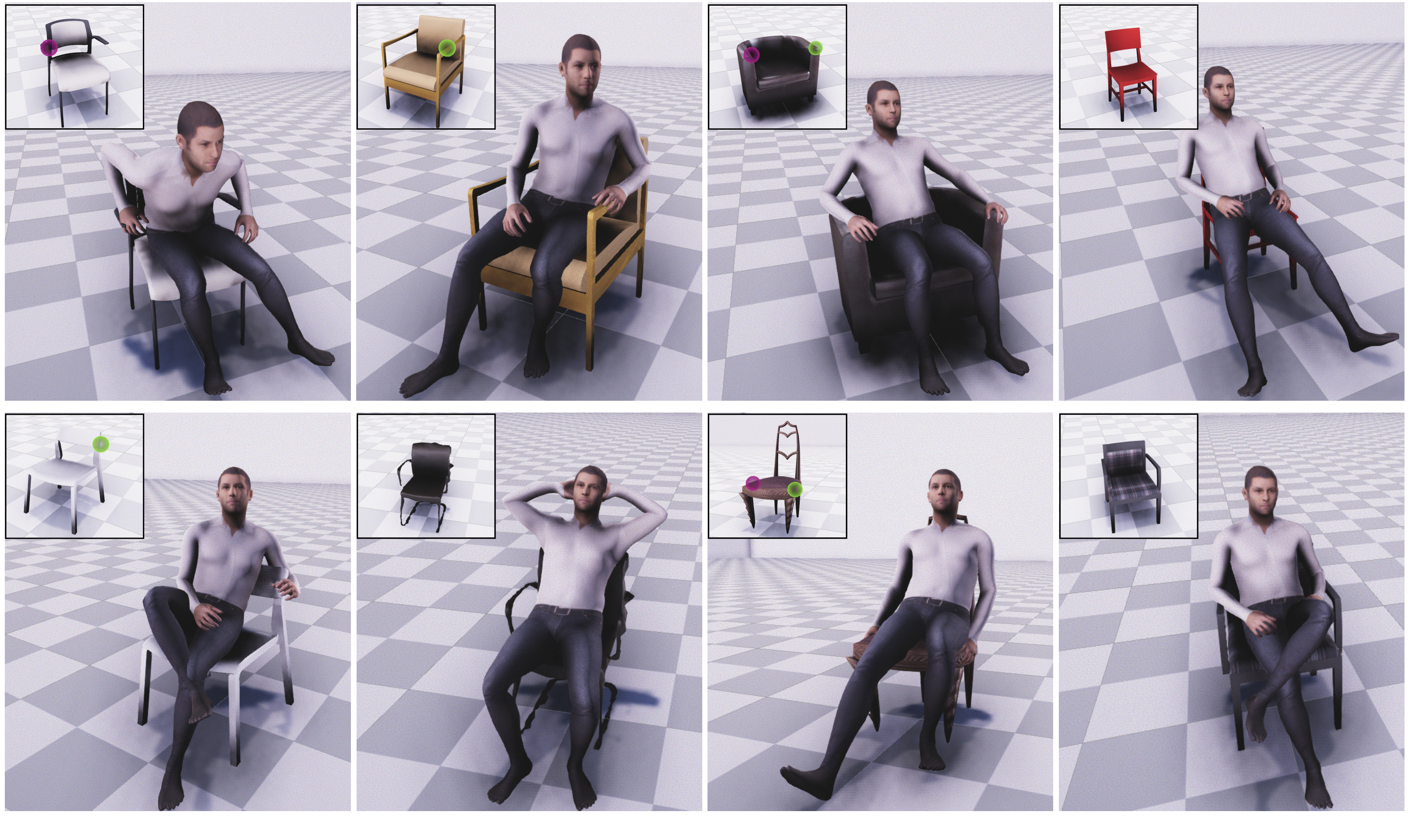}
\caption{We present \model. A dataset and model to synthesizes controllable, contact-based human-chair interactions.}
\squeezeup
\label{fig:teaser}
\end{center}
\end{figure}
\begin{abstract}
   Humans interact with an object in many different ways by making contact at different locations, creating a highly complex motion space that can be difficult to learn, particularly when synthesizing such human interactions in a controllable manner. Existing works on synthesizing human scene interaction focus on the high-level control of action but do not consider the fine-grained control of motion. In this work, we study the problem of synthesizing scene interactions conditioned on different contact positions on the object. As a testbed to investigate this new problem, we focus on human-chair interaction as one of the most common actions which exhibit large variability in terms of contacts. We propose a novel synthesis framework \model~that plans ahead the motion by predicting contact-aware control signals of the hands, which are then used to synthesize contact-conditioned interactions. Furthermore, we contribute a large human-chair interaction dataset with clean annotations, the \model~Dataset. Our method shows significant quantitative and qualitative improvements over existing methods for human-object interactions. More importantly, our method enables control of the motion through user-specified or automatically predicted contacts. Our dataset, model and code will be available at \cite{couch}. 

\end{abstract}

\section{Introduction}

% Cognitive agents that interact realistically in the world need to plan contact interactions to attain goals. We need research on this.
To synthesize realistic virtual humans which can achieve goals and act upon the environment, reasoning about the interactions and in turn contacts, is necessary. Reaching a goal, like sitting on a chair, is often preceded by intentional contact with the hand to support the body. In this work we investigate a motion synthesis method which exploits predictable contact to achieve more control and diversity over the animations. 

% PROBLEM: most works focus on humans in isolation or static humans. 
%Capturing human motion and 3D scene understanding play essential roles in various applications of virtual reality, video games, and robotics, but it remains challenge to the computer vision community to synthesize realistic human scene interactions. Although many algorithms are proposed for learning-based human synthesis~\cite{martinez,quaternet,Mao_2019_ICCV}, they focus on the motion of the human alone, without considering interaction with objects, or the scene.

% PROBLEM: What others can not do
Although most applications in VR/AR, digital content creation and robotics require synthesizing motion \emph{within the environment}, it is not considered in the majority of works in human motion synthesis~\cite{martinez,quaternet,Mao2019ICCV,aist}. Recent work does take the environment into account but is limited to synthesizing \emph{static poses}~\cite{PROX:2019,Zhang2020CVPR}.
Synthesising dynamic human motion coherent with the environment is a substantially harder task~\cite{nsm,samp,holden2020,starke2020,starke2021} and recent works show promising results. However, these methods do not reason about intentional contacts with the environment, and can not be controlled with user provided contacts. 

%Humans can interact with an object in the scene in many different ways, which are often associated with different approaches of contacting with the object. This creates highly complex motion manifold makes the learning difficult. It is even more challenging to learn to synthesize human interacting with an object in a controlled, desired way. Existing works on synthesizing human scene interaction focus on the high-level control of interacting with an particular object paying no attention to fine-grained control on the body limb movements.

% Solution and challenges
Thus, in this work, we investigate a new problem: synthesizing human motion conditioned on contact positions on the object to allow for controllable movement variations. As a testbed to investigate this new problem, we focus on human-chair interactions as one of the most common actions, which are of crucial importance for ergonomics, avatars in virtual reality or video game animation. Contact-driven motion synthesis is a more challenging learning problem compared to conditioning only on coarse object geometry~\cite{nsm,samp}.
%Human-chair interaction synthesis is already challenging, and controlling the motion with contacts is even more challenging. 
First, the human needs to approach the chair differently depending on the contacts, regardless of the starting position, walking around it if necessary. Second, a chair can be approached and contacted in many different ways; we can directly sit without using our hands, or we can first support the body weight using the left/right or both hands with different parts of the chair. Furthermore, individual styled free-interactions can be modelled such as leaning back, stretching legs, using hands to support the head, and so on. 

% motivation for our method 
Contact driven motion allows for providing more detailed instructions to the virtual human such as approaching to sit on the chair, while supporting the body with the left hand and placing it on the armrest, as illustrated in Figure \ref{fig:teaser}. Given the contact and the goal, the full-body needs to coordinate at run-time to achieve a plausible sequence of pose transitions. 
Intuitively, this emulates our planning of motion as real humans: we plan in terms of goals and intermediate object contacts to reach; the full-body then moves to follow such desired trajectories. %The $\textit{contact point}$ on the armrest needs to be associated together with the movement of the corresponding body limb (the left hand, in this case). Such information is processed while the character performs the interactions. Intuitively, this is very similar to our planning of motion as real humans: we plan in terms of the object contacts to reach; the full-body moves to follow such desired trajectories. 
% However, from the learning perspective, designing such detailed control signal can be difficult. It is shown in our experiments that conditioning current state-of-the-art neural auto-regressive models~\cite{nsm,samp} on the spatial location of contacts is not sufficient, and such signal is simply ignored by the network.
% \Seb{If saying this, also need to mention what enables this in this work: In previous sentence, you highlight the fine-grained control which is based on contacts, but then say that contacts are not sufficient (for existing works). What makes it sufficient in this work?}
%Synthesising such fine-grained interaction with a learned auto-regressive model is difficult. 
% Additionally the temporal aspect of the control signal has to be considered, since the body movement of different limbs can be desynchronized. \GPM{I don not understand the last sentence} \Seb{I think we should not mention this here as it may cause confusion for the reader. This temporal aspect is covered by phase, but which is not contribution of the work and would require more time putting into context. I think just mentioning it in related work / method should be fine? Otherwise, should be easier merged with the above sentence.}

% Overview of our method and components: mention real time, explain how contacts controls the style of interaction 
To this end, we propose \model, a method for controllable contact driven human-chair interactions, which is composed of two core components:
1) \textit{ControlNet} is responsible for motion planning by predicting the future control signal of the body limbs which guides the future body movement. Our spatial-temporal control signal consists of dynamic trajectories of the hands towards the contact points and the local phase, an auxiliary continuous variable that encodes the temporal information of a limb during a particular movement (e.g. the left hand reaching an armrest).
%At the beginning of a key movement, the phase has value 0, and reaches to 1 when the movement is complete.
2) \textit{PoseNet} conditions on the predicted control signal to synthesise motion that follows the dynamic trajectories, ensuring the contact point is reached.
At runtime, \model~can operate in two modes. First, in an interactive mode where the user specifies the desired contact points on the target object. Second, in a generative mode where \model~can automatically sample diverse intentional contacts on the object with a novel neural network called \textit{ContactNet}.
Training and evaluating \model~calls for a dataset of rich and accurate human chair interactions. Existing interaction 3D datasets~\cite{nsm} are captured with Inertial Sensors, and hence do not capture the real body motion and the contacts with the real chair geometry -- instead synthetic chairs are fit to the avatar as post-process in those works. Hence, to jointly capture real human-chair interactions with fine-grained contacts, we fit the SMPL model \cite{smpl} and scanned chair models to data obtained from multiple Kinects and IMUs. The dataset (the \model~dataset, Table \ref{table:dataset}) consists of 3 hours (over 500 sequences) of motion capture (MoCap) on human-chair interactions. Moreover it features multiple subjects, accurately captured contacts with registered chairs, and annotation on the type of hand contact. Our experiments demonstrate that \model~runs in real-time at 30 fps, \model~generalizes across chairs of varied geometry, and different starting positions relative to the chair. Compared to SoTA models (trained on the same data) adapted to incorporate contacts, our method significantly outperforms them in terms of control by improving the average contact distance by 55$\%$.

The contributions of our work can be summarized as follows:

\begin{itemize}
    \item We propose \model, the first method for synthesizing controllable contact-based human-chair interactions. Given the same input control, the \model~model can achieve diverse sitting motions. By specifying different control signals, the user enables control over the style of interaction with the object. Results show our method outperforms the state of the art both qualitatively and quantitatively. 
    \item To train \model, we captured a large-scale MoCap dataset consisting of 3 hours (over 500 sequences) of human interacting with chairs different styles of sitting and free movements. The dataset features multiple subject, real chair geometry, accurately annotated hand contacts, and RGB-D images. 
    %Each action class has different action modes which provides diversity.
    \item To stimulate further research in controllable synthesis of human motion, we will release the \model~model and dataset.
\end{itemize}

% \begin{figure*}[t!]
%     \includegraphics[width=\textwidth]{figures/architecture.png}
%     \caption{\model comprises of two modules, `Probabilistic Trajectory Prediction' and `Trajectory Conditioned Pose Prediction'. Former is a generative network that takes the goal $\goal$, past trajectories $\hist{\trajectory}$, phase $\phi_{t-1}$, chair $\scene$ and sampling variable $\vect{z}_t$ as input and predicts future trajectories $\future{\trajectory}$. $\vect{z}_t$ is sampled from $\set{N}(0, I)$ at inference and encoded using a network (shown in dotted lines) during training.
%     The trajectory conditioned pose prediction network takes the chair $\scene$, goal $\goal$, trajectories predicted by the previous network $\future{\trajectory}$ and previous pose $\pose_{t-1}$ as input and predicts future pose $\pose_t$.}
%     \label{fig3:architecture}
% \end{figure*}

\section{Related Work}
\subsubsection*{Scene agnostic human motion prediction.} 
Synthesizing realistic human motion has drawn much attention from the computer vision and graphics communities. However, many methods do not take the scene into account. Existing methods on short ($\sim$1 sec)~\cite{martinez,Gui2018ECCV,quaternet,Cui2020CVPR,Xueccv20,vred,Aksan2019ICCV,yuan2020dlow,moglow20} and long ($>$1 min)~\cite{motion2016holden,ghosh2017learning} term 3D human motion prediction aim to produce realistic-looking human motion (typically walking and its variants). There also exists work on conditional motion generation based on music~\cite{aist}. These methods have two major limitations, i) except for work that use generative models~\cite{yu2020character,BMVC2017119,Gui2018adversarial,Aliakbarian2020CVPR,Hernandez2019ICCV}, these methods are largely deterministic and cannot be used to generate diverse motions and ii) this body of work is unfortunately agnostic to scene geometry~\cite{Li2020CVPR,Mao2019ICCV}, which is critical to model human scene interactions.
Our method on the other hand can generate \emph{realistic motion and interactions}, taking into account the 3D scene.

\subsubsection*{Affordance and Static Scene Interactions.} Although the focus of our work is to model human-scene interactions over time, we find the works predicting static affordances in a 3D scene~\cite{Li2019CVPR} relevant.
This branch of work aims at predicting static humans in a scene~\cite{Zhang2020CVPR,PLACE:3DV:2020,PROX:2019} that satisfies the scene constraints. More recently there have been attempts to model fine-grained interactions (contacts) between the hand and objects~\cite{Corona2020CVPR,GRAB:2020,Brahmbhatt2019CVPR,GrapingField:3DV:2020}. 

The aforementioned methods focus on predicting static humans / human poses that satisfy the scene constraints in case of affordances or grasping an object in case of hand-object interactions. But these methods cannot produce a full sequence of human motion and interaction with the scene. Ours is the first approach that can model fine-grained interactions (contacts) between an object in the scene and the human. 
%Our approach can not only predict the contacts, but can also predict natural human motion to satisfy these contacts.

\subsubsection*{Dynamic Scene Interactions.} Although various algorithms have been proposed for scene-agnostic motion prediction, affordance prediction as well as the synthesis of static human-scene interaction, generating dynamic human-scene interactions is less explored. Recent advances include predicting human motion from scene images~\cite{caoHMP2020}, and using a semantic graph and RNN to predict human and object movements~\cite{corona2019}. More recently, Wang et al.~\cite{wang2020} introduce a hierarchical framework that generates `in-between' locations and poses on the scene and interpolates between the goal poses. However, it requires a carefully tuned post-optimization step over the full motion synthesis to solve the discontinuity of motion between sub-goals and to achieve robust foot contacts with the scene. Alternatively, Chao et al.~\cite{learningtosit} use a reinforcement learning based approach by training a meta controller to coordinate sub-controllers to complete a sitting task.
An important category of human-scene interaction involves performing locomotion on uneven terrains. The Phase-functioned Neural Network~\cite{pfnn} first introduced the use of external phase variables to represent the state of the motion cycle. Zhang et al.~\cite{mann} applies same concept for quadruped motion and further incorporates a gating network that segments the locomotion modes based on foot velocities. Both works show impressive results thanks to the mixture of experts~\cite{moe} styled architectures.

The most relevant work to us, are the Neural State Machine (NSM)~\cite{nsm} and SAMP~\cite{samp}. While NSM is a powerful method and models human-scene interactions such as sitting, carrying boxes and opening doors, it does not generate motion variations for the same task and object geometry. SAMP predicts diverse goal locations in the scene for the virtual human, which is then used to drive the motion generation. Our work takes inspiration from these works, but it is demonstrated qualitatively and quantitatively from our experiments that neither of the work enables control over the style of interaction (Section \ref{subsec:control}). Our work focus on controllable, fine-grained interactions given on contacts on the object. To the best of our knowledge, no previous work has tackled the problem of generating controllable human-chair interactions.
% We approach such challenge via contact-based motion planning and contact-conditioned motion synthesis.
%  By doing so, the trajectories can subsequently be used as a part of the control signal to synthesis the full body motion. 

\section{The \model~Dataset}
\begin{figure}[t!]
\begin{center}
\fbox{\includegraphics[width=\linewidth]{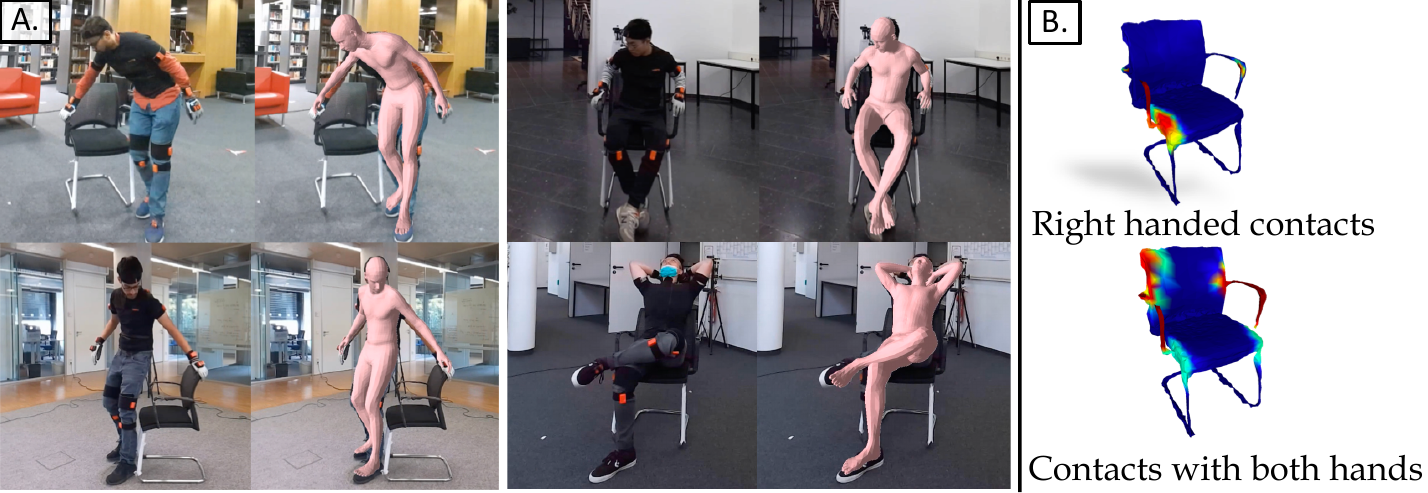}}
\caption{(A) The \model~dataset captures a diverse range chair interaction with an emphasis on hand contacts. It consists of RGB-D images, corresponding MoCap in the SMPL \cite{smpl} format, chair geometry and annotations on types of hand interaction.
% We capture six subject within four indoor scenes. The left column shows examples of different fitted SMPL model of which subject contacts with the chair with the left hand, right hand, and both hands, respectively, while the right column illustrates top clusters of contact heatmaps on a chair, the clusters represents subject contacts the chair with the left hand, right hand, or both hands.
(B) \model~dataset captures natural modes of interactions with a chair, as demonstrated by the heatmaps of contact clusters. Most common contacts while sitting, include right hand support or both hands.
}
\label{fig:control}
\end{center}
\end{figure}

Large scale datasets of human motion such as AMASS~\cite{amass} and H3.6m~\cite{h36mpami} have allowed us to build models of human motion. Unfortunately these datasets only contain sequences of 3D poses but no information about the 3D environment, making these datasets unsuitable for learning human interactions.
On the other hand, datasets containing human-object interactions are either restricted to just hands~\cite{Brahmbhatt2019CVPR}, contain only static human poses~\cite{GRAB:2020,PROX:2019} without any motion, or have little variation in motion~\cite{nsm}.

We present a multi-subject dataset of human chair interactions. Our dataset consists of 6 different subjects interacting with chairs with over 500 motion sequences. We collect our dataset using 17 wearable Inertial Measurement Units (XSens)~\cite{xsens}, from which we obtain high-quality pose sequences in SMPL~\cite{smpl} format using Unity~\cite{unity}. The total capture length is~3 hours. 

\begin{table}[t]
    \small
    	\caption{Comparison with existing motion capture datasets for human chair interactions. The \model~dataset features registered real chairs models, multiple subject, and RGB-D data. The types of hand contact are also annotated.} 
	\centering
	\begin{tabular}[b]{c|ccc}
		\hline
		{Features} & {NSM\cite{nsm}} & {SAMP\cite{samp}}  & {Ours}  \\
        \hline
        $\textrm{Real Objects}$ & \xmark & \cmark & \cmark\\
        $\textrm{Multiple Subjects}$ & \xmark & \xmark & \cmark \\
        $\textrm{Contact Types}$ & \xmark & \xmark & \cmark \\
        $\textrm{RGB-D}$ & \xmark & \xmark & \cmark \\
        % $\textrm{Multi-Action Modes}$ & \xmark & \xmark & \cmark \\
    	\hline
	\end{tabular}
    \label{table:dataset}
\end{table}
% \smallsqueezeup

% \begin{table}[t]
%     \small
%     	\caption{Comparison with existing motion capture datasets for human chair interactions. The \model dataset features registered real chairs models, multiple subject, and RGB-D data. The types of hand contact are also annotated.} 
% 	\centering
% 	\begin{tabular}[b]{c|ccc}
% 		\hline
% 		{Features} & {Starke et al.,} & {Hassan et al.,}  & {Ours}  \\
%         \hline
%         $\textrm{Real Objects}$ & \color{red}\xmark & \color{green}\cmark & \color{green}\cmark\\
%         $\textrm{Multiple Subjects}$ & \color{red}\xmark & \color{red}\xmark& \color{green}\cmark \\
%         $\textrm{Contact Types}$ & \color{red}\xmark & \color{red}\xmark & \color{green}\cmark \\
%         $\textrm{RGB-D}$ & \color{red}\xmark & \color{red}\xmark & \color{green}\cmark\\
%         % $\textrm{Multi-Action Modes}$ & \xmark & \xmark & \cmark \\
%     	\hline
% 	\end{tabular}
%     \label{table:dataset}
% \end{table}

Motion capture with marker-based capture systems is restrictive to capturing human-object interactions because markers often get occluded during the interactions leading to inaccurate tracking.
IMU-based systems are prevalent for large-scale motion capture, however, the error from its calibration can lower the accuracy of the motion.
We propose to combine IMUs with Kinect-based capture system as an efficient trade-off between scalability and accuracy.
Our capture system is lightweight and can be generalized to capture many human scene interactions.
We use the SMPL registration method similar to~\cite{bhatnagar2020ipnet,rong2021frankmocap,alldieck19cvpr} to obtain SMPL fits for our data. The dataset is captured in four different indoor scenes. The average fitting error for the SMPL human model, and the chair scans to the point clouds from the Kinects are 3.12 cm and 1.70 cm, respectively (in Chamfer distance). More details about data capture can be found in supp. mat.
% The motions vary in starting point and type of interaction. Additionally, to obtain variation in chair geometry, we augment the dataset by fitting different 3D chair models from ShapeNet~\cite{shapenet}. 

\myparagraph{Diversity on Starting Points and Styles.} We capture people approaching the chairs from different starting points surrounding the chairs. Each subject then performs different styles of interactions with the chairs during sitting. This includes, one hand touching the arm of the chair, both hands touching the armrests of the chair, one hand touching the sitting plane of the chair before sitting down, and no hand contacts. It also includes free interactions such as crossing legs or leaning forward and backward on the chairs. To ensure the naturalness of motion, each subject is only provided with high-level instruction before capturing each sequence and was asked to perform their styles freely. Annotations of the direction of the starting points relative to the chair as well as the type of hand contact are included in the dataset.

\myparagraph{Objects.} Our dataset contains three different chair models that vary in terms of their shapes, as well as a sofa. The objects are 3D scanned \cite{agrisoft,treedys} before registering into the Kinect captured point clouds. To generalize the synthesized motion to unseen objects, we perform data augmentation as in \cite{nsm}. 

\myparagraph{Contacts.} Studying contact-conditioned interaction calls for accurate contacts to be annotated in the dataset. Since we capture both the body motion and the object pose, it is possible to capture contacts between the body and the object. We detect the contacts of five key joints of the virtual human skeleton, which are the pelvis, hands, and feet. We then augment our data by randomly switching or scaling the object at each frame. The data augmentation is performed on 30 instances from ShapeNet \cite{shapenet} over categories of straight chairs, chairs with arms, and sofas. At every frame, we project the contacts detected from the ground truth data to the new object, and apply full-body inverse kinematics to recompute the pose such that the contacts are consistent, keeping the original context of the motion. 
% We end up with ~24k training samples, where the threshold for contact detection is set to 0.05 cm.
% Studying contact-conditioned interaction calls for accurate contacts to be annotated in the dataset.

\begin{figure}[t!]
\begin{center}
\includegraphics[width=1.0\linewidth]{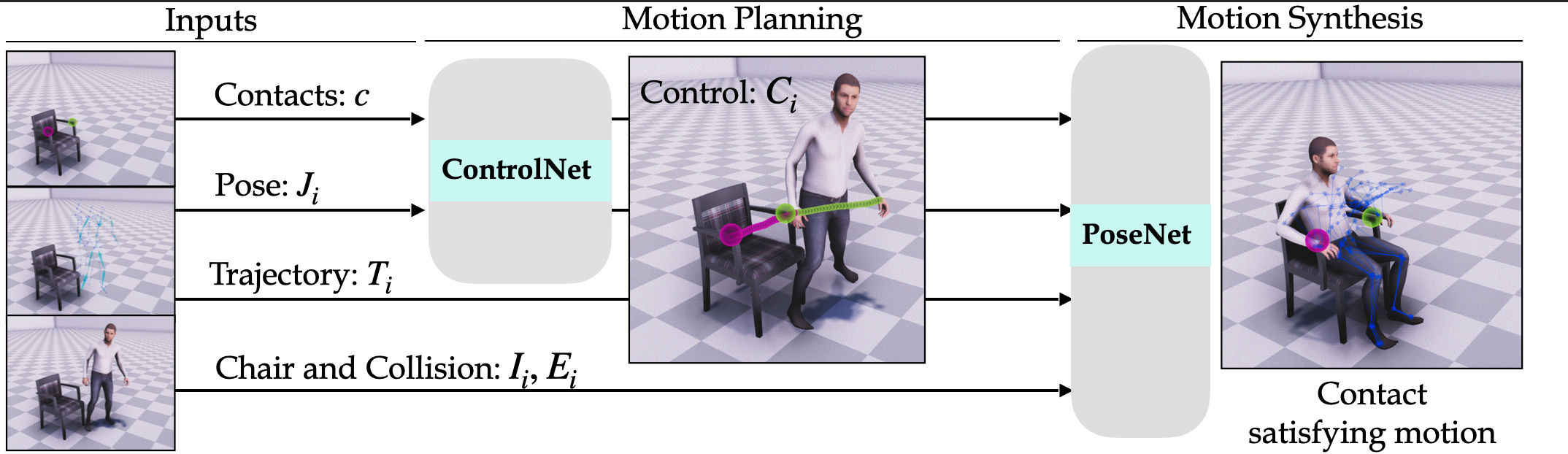}
\caption{Given user specified or model generated contacts, our proposed method which consists of the ControlNet and the PoseNet, auto-regressively synthesizes contact satisfying motion.}
\label{fig:controlnet}
\end{center}
\end{figure}

\section{Method}
We address the problem of synthesising 3D human motion that is natural and satisfies environmental geometry constraints and user-defined contacts with the chair. \model~allows fine-grained control over how the human interacts with the chair. At run-time, our model operates in two modes. First, a generative mode where \model~can automatically sample diverse intentional contacts on the object with our proposed generative model. Second, an interactive mode where the user specifies the desired contact points on the target object. 
\\
The input to our method is the current character pose, the target chair geometry as well the target contacts for the hands that need to be met. Our method takes these inputs and predicts the future poses that satisfy the desired contacts auto-regressively.

\subsection{Key Insights}
Synthesising natural human motion subject to environmental constraints is a challenging task \cite{nsm,samp,pfnn}, particularly when also satisfying a set of desired contacts.
% In this work, we address an even more challenging problem as our synthesised motion must additionally satisfy the desired contacts.
To this end, we first divide our motion synthesis task into \emph{motion planning} and \emph{motion prediction}. We derive our intuition from the way humans execute complex interactions e.g., to sit on the chair, we first prepare a mental model of how we will sit (place a hand on the arm-rest and sit, place a hand on the sitting plane and sit or just sit without using the hands etc.) and then we move our bodies accordingly. We propose two neural networks \emph{ControlNet} $f^\text{CN}(\cdot)$, and \emph{PoseNet} $f^\text{PN}(\cdot)$, for motion planning and motion prediction respectively.
\\
Furthermore, we observe that it is useful to perform detailed hand motion planning only when we are close to the chair right before sitting and not when we are far off. Thus, we decompose the motion synthesis into \emph{approaching} and \emph{sitting}. The \emph{approaching} motion can be generated directly with \emph{PoseNet} but both networks are required for \emph{sitting}, \emph{ControlNet} and \emph{PoseNet}, for generating the sitting motion that satisfies the given contacts.

% \paragraph{Encoding contacts} The desired hand contacts $\contact \in \mathbb{R}^{2 \times 3}$ are represented as 3D points on the surface of the input chair where the hands should reach.

\subsection{Motion Planning with ControlNet}
\label{sec:controlnet}
ControlNet is the core of our method and plays an important role in motion planning, that is predicting the future control signals of the key joints which are used to guide the body motions. At a high level, the contact-aware control signal contains the local phases and the future locations of the key joints (in our case, the two hands). The local phase is an auxiliary variable that encodes the temporal alignment of motion for each of the hands and prepares for a future contact to be made. When the virtual human is ready to make contact with the chair, and at the beginning of the hand movement, the local phase is equal to 0, and it gradually reaches the value 1 as the hand comes closer to the contact. The hand trajectory, on the other hand, encodes the spatial relationship between the hand joint and the given contact location.

More formally, we define our spatial-temporal control signal at frame $i+1$ to be $\control_{i+1}^+=\{\handtraj_{i+1}^{+}, \localphase_{i+1}^{+}\}$, where  $\handtraj_{i+1}^+ \in \mathbb{R}^{2  \times 3\times \tau^+}$ represents the future position of the two hand joints relative to their corresponding desired contact point $\contact \in \mathbb{R}^{2 \times 3}$, and their local phases are represented by $\localphase_{i}^+ \in \mathbb{R}^{2 \times \tau^+}$. 
We predict the control signal for $\tau^+=7$ time stamps sampled uniformly between [0, 1] second window centered at frame $i+1$.
\\
We use an LSTM based $f^\text{CN}(\cdot)$ to predict the control signal,
\begin{equation}
    \control_{i+1}^+ = f^\text{CN}(\handtrajinterp_{i},  \localphase_{i}),
\end{equation}
where $\handtrajinterp_{i} \in \mathbb{R}^{2 \times 3\times \tau^+ }$ denote $\tau^+$ points interpolated uniformly on the straight line from the current hand locations to their desired contact locations $\contact$. Intuitively, these interpolated positions encourages the ControlNet to predict future hand trajectories that always reach the given contacts. $\localphase_{i} \in \mathbb{R}^{2 \times \tau}$ denotes the local phases of the hands over $\tau = 13$ frames sampled uniformly between the [-1, 1] second window centered at frame $i$.

The ControlNet is trained to minimize the following MSE loss on the future hand trajectories and the local phase, which is formulated as follows:
\begin{equation}
\begin{split}
    L_{\mathrm{control}} = \lambda_{1}\|\handtraj_{i+1}^+ - \hat{\handtraj}_{i+1}^+  \|_{2}^{2} + \lambda_{2}\|\localphase_{i+1}^+ - \hat{\localphase}_{i+1}^+ \|_{2}^{2} +  \lambda_{3}L_{\mathrm{reg}}.
 \end{split}
\label{eq:controlnet}
\end{equation}
Here, $\handtraj_{i+1}^+, \localphase_{i+1}^+$ are the network predicted future trajectories and local phases. $\hat{\handtraj}_{i+1}^+, \hat{\localphase}_{i+1}^+$ are the corresponding GT. We also introduce an additional regularization term $L_{\mathrm{reg}}=\|\handtraj_{i+1}^+ - \handtrajinterp_{i}\|_2$. Please see supplementary for implementation details regarding the network architectures and training.

\subsection{Motion Synthesis with PoseNet}
\label{sec:posenet}
ControlNet generates important signals that guide the motion of the person such that user-defined contacts are satisfied. To this end, we train PoseNet $f^\text{PN}(\cdot)$, that takes as input the control signals predicted by the ControlNet along with the 3D scene and motion in the past and predicts full body motion.

\begin{equation}
    \pose_{i+1}, \traj_{i+1}^+, \goal_{i+1}^+, \globalphase_{i+1}, \futureposeposition, \goaltraj_{i+1}, \binarycontact = f^\text{PN}( \control_{i}^+, \pose_{i}, \traj_{i}, \goal_{i}, \scene_i ,\environment_{i}, \globalphase_i),
\end{equation}
where $\control_{i}^+$ is the control signal generated by the \emph{ControlNet}. We represent the current state of motion for the human model: $\pose_{i}= ( \poseposition, \posevelocity, \poserotation)$ contains root relative position $\poseposition\in R ^{j\times 3}$, rotation $\posevelocity\in R ^{j \times 6}$ and velocity $\poserotation\in R ^{j \times 3}$ of each joint at frame $i$. We use $j=22$ joints for our human model. $\mat{T}_{i} = (\trajpos, \trajdir, \trajact)$ contains the root positions $\trajpos\in R ^{\tau \times 3}$ and rotation $\trajdir \in R ^{\tau \times 6}$ for $\tau=13$ frames sampled uniformly between the [-1, 1] second window centered at frame $i$.
$\trajact \in R ^{\tau \times 3}$ are the soft labels which describe current action over ours three action classes, namely, idle, walk, and sit. Inspired by Starke et al.,~\cite{nsm}, we also use intermediate goals $\goal_i = (\goalpos, \goaldir, \goalact)$, where $\goalpos \in R ^{\tau \times 3}$, $\goaldir \in R ^{\tau \times 6}$ are the goal positions and orientations at frame $i$. $\goalact \in R ^{\tau \times 3}$ are the one-hot labels describing the intended goal action.
\\
To accurately capture the spatial relation between the person and the chair, we voxelize the chair into an $8\times8\times8$ grid and store at each voxel its occupancy ($\mathbb{R}$) and the relative vector between the root joint of the person and the voxel ($\mathbb{R}^3$). This allows us to reason about the distance between the person and different parts of the chair. We flatten this grid to obtain our chair encoding
$\scene_i \in \mathbb{R}^{2048}$ at time-step $i$.
\\
In order to explicitly reason about the collisions of the person with the chair, we voxelize the region around the person into a cylindrical ego-centric grid and store the occupancies corresponding to the chair (if it is inside the grid). We flatten the occupancy feature to obtain $\environment_{i} \in \mathbb{R}^{1408}$. 
It is important to note that although $\scene_i$ and $\environment_{i}$ are scene encodings that serve different purposes. $\scene_{i}$ is chair-centric and entails information about how far is the person from the chair and the geometry of the chair, while $\environment_{i}$ is ego-centric and detects collisions in the surrounding of the human model. In addition, we also introduce an auxiliary variable $\globalphase \in [0,1]$ as in~\cite{quaternet,pfnn}, which encodes the global phase of the motion. When approaching the goal, the represents the timing within a walking cycle, for sitting the phase equals $0$ when the person is still standing and reaches $1$ when the person has sat. \\
The components of the output of the network differs from the input to a small extend by additionally predicting $\futureposeposition$ are the joint positions relative to future root 1 second ahead. To ensure the human model can reach the chair, we introduce the goal-relative root trajectory $\goaltraj_{i+1} = \{\goaltrajpos, \goaltrajdir\}$ which include the root positions and forward directions relative to the chair of frame $i+1$. The rest of the components remain consistent with the input include the the future pose $\pose_{i+1}$, future root trajectory $\traj_{i+1}^+$, the future intermediate goals $\goal_{i+1}^+$, and the future global phase $\globalphase_{i+1}$. The PoseNet $f^\text{PN}(\cdot)$ adopts a mixture-of-experts \cite{pfnn,samp,nsm,starke2020} and is trained to minimize the standard MSE loss.

\subsection{Contact Generation with ContactNet}
\label{sec:contactnet}
From the user's perspective, it is useful to automatically generate plausible contact points on any given chairs. To this reason, we propose \emph{ContactNet}. The network adopts a conditional variational auto-encoder\cite{cvae} architecture (cVAE) which encodes the chair geometry $\scene$ introduced in Section \ref{sec:posenet} and the contact positions $\contact \in R^{2 \times 3}$ to a latent vector $\latent$. The decoder of the network then reconstructs the hand contacts $\hat{\contact} \in R^{2 \times 3}$. Note, the position of each voxel in the scene representation $\scene$ in this case is computed relative to the center of the chair instead of the character's root. During training, the network is trained to minimize the following loss,
\begin{equation}
\begin{split}
    L_{\mathrm{contact}} = \|\hat{\contact} - \contact\|_{2}^{2}  + \beta KL(q(\latent | \contact, \scene )\| p(\latent)),
 \end{split}
\label{eq:contactnet}
\end{equation}
where $KL$ denotes the Kullback-Leibler divergence. During inference, given the scene representation  $\scene$ of a novel chair, we sample the latent vector $\latent$ from the uniform Gaussian distribution $\set{N}(\vect{0},\vect{I})$, and use the decoder to generate plausible hand contacts $\contact \in \mathbb{R}^{2\times3}$. 

\subsection{Decomposition of \emph{Approaching} and \emph{Sitting} Motion}
Detailed hand motion planning is only required when the human model is close enough to the chair right before sitting as sitting requires synthesizing more precise full-body motion, especially for the hands, such that the person makes the desired contacts and sits on the chair. For this reason, we decompose our synthesis into approaching and sitting by only activating the ControlNet during the sitting. When the ControlNet is deactivated the control signal or when a ``no contact" signal is present the control signal for the corresponding hand is zeroed.

\section{Evaluation}
Studying contact-conditioned interaction with chairs requires accurately labelled contacts and a diverse range of styled interactions. The \model~dataset is captured to meet such needs. We evaluate our contact constrained motion synthesis method on the \model~dataset qualitatively and quantitatively. Our method is the first approach that allows the user to explicitly define how the person should contact the chair and we generate natural and diverse motions satisfying these contacts. As such we evaluate our method on three axis, (i) accuracy in reaching the contacts, (ii) diversity and (iii) naturalness of the synthesised motion. For qualitative results, we highly encourage the readers to see our supplementary video. It can be seen that our method can generate diverse and natural motions while reaching the user-specified contacts. We quantitatively evaluate the accuracy of contacts and motion diversity on a total of 120 testing sequences on six subject-specific models trained on corresponding subsets of our \model~dataset. Note that we evaluate raw synthesized motion without post-processing.

\subsection{Baselines}
To our best knowledge, the most related work to ours are the Neural State Machine (NSM) \cite{nsm} and the SAMP \cite{samp} since they both synthesize human-scene interactions. However, neither of the methods allows the use of fine-grained control over how the interaction should take place. We adapt these baselines for our task by additionally conditioning on the contact positions and refer to these new baselines as NSM+Control and SAMP+Control. Quantitative results are reported for both the original baselines and their adapted version. For each of the methods, we train subject-specific models with the corresponding subset of our \model~dataset using the code provided by the authors. Our experiments, detailed below, show that naively providing contacts as input to existing motion synthesis approaches does not ensure that the generated motion satisfies the contacts. Our method, on the other hand, does not suffer from this limitation.

\label{subsec:baselines}
\begin{figure}[t!]
\begin{center}
\includegraphics[width=0.8\linewidth]{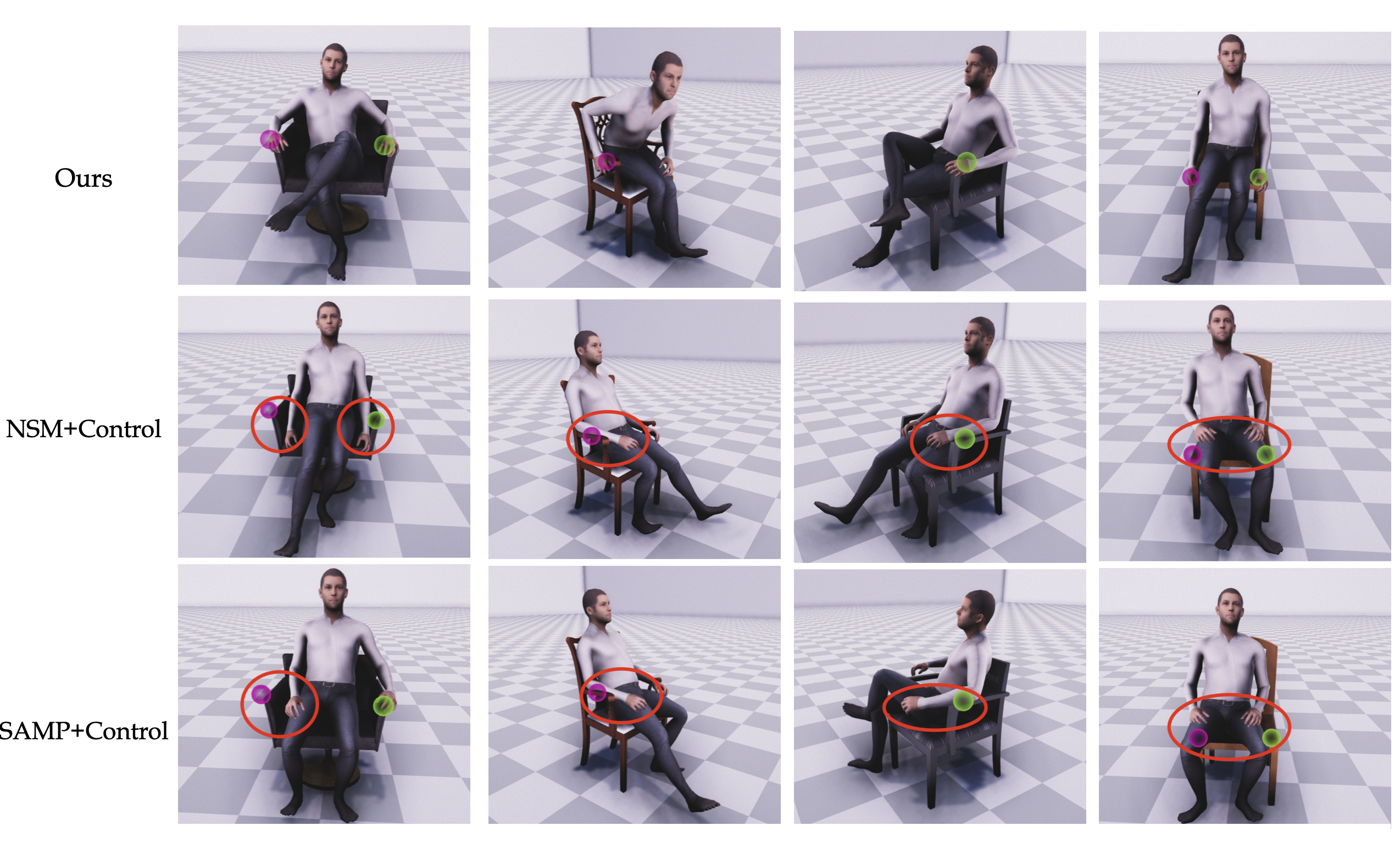}
\caption{We demonstrate qualitatively and quantitatively (Tab.~\ref{table:control}) that motion generated by our approach satisfies the contacts much better than the baselines, NSM+Control \cite{nsm} and SAMP+Control \cite{samp}.}
\label{fig:control}
\end{center}
\end{figure}

\begin{figure}[t!]
\begin{center}
\includegraphics[width=\linewidth]{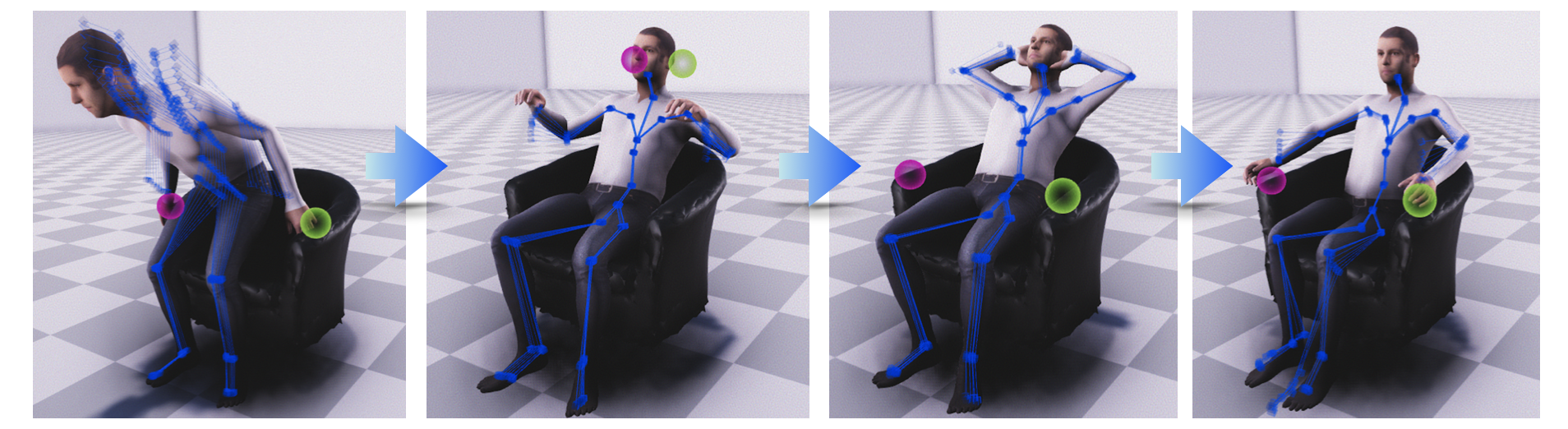}
\caption{\model~can also be extended by specifying a series of contacts for to automatically synthesize more complex interactions. The past poses are indicated by blue skeletons.}
\label{fig:serial}
\end{center}
\end{figure}

%%%%%%%%%%%%%%%%%%%%%%%%%%%%%%%% Contact prediction %%%%%%%%%%%%%%%%%%%%%%%%%%
\subsection{Evaluation on Control} 
\label{subsec:control}

In order to evaluate how well our synthesised motion meets the given contacts, we report the \emph{average contact error (ACE)} as the mean squared error between the predicted hand contact and the corresponding given contact. We use the closest position of the predicted hand motion to the given contact as our hand contact.
\\
Since ACE might be susceptible to outliers and inspired by the literature on object detection~\cite{fpn,fastrcnn,fasterrcnn,detr}, we also report \emph{average contact precision} (AP@k), where we consider a contact as correctly predicted if it is closer than $k$ cm. 

We compare our method with NSM+Control and SAMP+Control in Table \ref{table:control}.
It can be observed that \model~outperforms prior methods by a significant margin. Prior methods are trained to condition on the contact positions, however it is found (Figure \ref{fig:control}) to be not sufficient as the contact input can be easily ignored during auto-regressive prediction. As a result the contact constraints are often not met. This highlights the importance of motion planning in form of trajectory predictors in order to reach the desired contacts. Our ControlNet provides valuable information on how to synthesise motion such that the given contacts are satisfied. Our motion prediction network PoseNet uses these control signals to generate contact constrained motions.
\tinysqueezeup
\begin{table}[h]
    \small
    	\caption{Evaluation on degree of control. \model~is shown to be more controllable compared to the baseline methods. The distance from given contact points and the joint position are measured. The success rate of control is also reported.}
	\centering
	\resizebox{\columnwidth}{!}{%
	\begin{tabular}{cccccc}
		\hline
		{Method} & {$\text{Distance to Contact}^{\downarrow}$} & {$\text{AP@ 3 cm}^{\uparrow}$} & {$\text{AP@ 5 cm}^{\uparrow}$} & {$\text{AP@ 7 cm}^{\uparrow}$} & {$\text{AP@ 9 cm}^{\uparrow}$}\\
		\hline
		NSM \cite{nsm}   &  10.69   & 15.52 & 38.20  & 46.05 & 56.61\\
    	SAMP \cite{samp} &  11.96    &  6.54 & 14.57 & 20.94  &  50.83\\
        \hline
		NSM+Control \cite{nsm}   &  10.52   &  17.46 & 35.7  & 48.4 & 57.93\\
    	SAMP+Control \cite{samp} &  12.09   &  7.20  & 15.2  &  23.2 & 48.80\\

		Ours  & \bf{4.73}   & \bf{47.97} & \bf{78.86} & \bf{87.8} & \bf{91.87}\\
		\hline
	\end{tabular}
	}
    \label{table:control}
\end{table}
\smallsqueezeup

% The PoseNet which conditions on the signal is able to perceive better on where and whether the contact should take place. 

%%%%%%%%%%%%%%%%%%%%%%%%%%%%%%%% Motion Diversity %%%%%%%%%%%%%%%%%%%%%%%%%%
\subsection{Evaluation on Motion Diversity}
\label{subsec:diversity}

Diversity is an essential element for our motion synthesis, since a chair can be approached and interacted with in different ways. To quantify diversity, we evaluate using the Average Pairwise Distance (APD) \cite{yuan2020dlow,Zhang2020CVPR,samp} on the synthesized pose features of the virtual human $\pose_{i} = ( \poseposition, \posevelocity, \poserotation)$. 
defined as:
\begin{equation}
\begin{split}
APD = \frac{1}{N(N-1)}\sum_{i=1}^{N}\sum_{j\neq i }^{N}D(\pose_{i}', \pose_{j}'),
\end{split}
\label{eqn:apd1}
\end{equation}
where $N$ is the total number of frames in all the testing sequences.
Note that for evaluation, the virtual human is initialized at different starting points and is instructed to approach and sit on randomly selected chairs with randomly sampled contact points from the dataset, and motion is synthesized for 16 seconds for each sequence. We compare the diversity of synthesized motion in Table \ref{table:apd1} and it can be seen that using explicit contacts allows our method to generate more varied motion.

\tinysqueezeup
\begin{table}[h]
    \small
    	\caption{Evaluation on the diversity of the synthesized motion. APD is measured for segmented motion of approaching and sitting. Our approach attains the best score compared to the baselines.}
	\centering
	\begin{tabular}{ccc}
		\hline
		{Method} & {$\text{Approach}$} & {$\text{Sit}$} \\
		\hline
		NSM \cite{nsm}   &  5.15   & 5.76\\
    	SAMP \cite{samp} & 5.34   &  5.81 \\
        \hline
		NSM+Control \cite{nsm} & 5.07 &  5.80 \\
    	SAMP+Control \cite{samp} &  5.21  & 5.88 \\
		Ours  &  {\bf 5.55}   & {\bf 6.02} \\
		\hline
		Ground Truth  & 5.69  & 6.30 \\
		\hline
	\end{tabular}
	%}
    \label{table:apd1}
\end{table}
\smallsqueezeup

\subsection{Controlling with a series of Contacts.}
A useful application of our approach is to automatically generate a motion sequence with a series of desired contacts in the context of animation, character control, when executing a set of complex actions. For instance, the person can be instructed to first sit with their hands on the armrest, then lift the arms to support the head before bringing the hands back to the armrest, see Figure \ref{fig:serial}) and the supplementary video. Our approach can be adapted for this task by iteratively providing the new goal locations for the hands as input after the present locations are reached.

% Furthermore, the control mechanism can be extended such that a serial of control signals that regard to different positions can be provided to the virtual human at different times. This enables the automatic generation of more complex motion. As shown in Figure \ref{fig:serial}), the virtual human is initially instructed to place its hands on the armrests of the sofa before a change in the control signal to lift them to supporting the head. This motion is then followed by resting the hands back on the armrests. During this automatic synthesis, the control signal evolves to be relative to the next location once the previous location is reached.

%%%%%%%%%%%%%%%%%%%%%%%%%%%%%%%% Diversity %%%%%%%%%%%%%%%%%%%%%%%%%%
% \myparagraph{Diversity.} The quantitative evaluation on the diversity of synthesized motion is shown in Table \ref{table:apd1}. Where each synthesized motion is split between approaching and sitting, and the diversity for each action class is evaluated separately. For the metric our joint positions, \model generates more diverse motion with an APD of XXXX metres, versus XXXX metres for NSM and XXXX metres for SAMP. Similarly, \model outperforms the baselines with an APD of XXXX degrees, versus XXXX degrees for NSM and XXXX degrees for SAMP. It is also shown qualitatively in Figure \ref{fig:teaser}, that \model is capable of synthesizing diverse interactions that satisfy the contact constraints. 

\subsection{Contact Prediction on Novel Shapes}
\label{subsec:contact}  
\begin{figure}[t!]
\begin{center}
\includegraphics[width=0.8\linewidth]{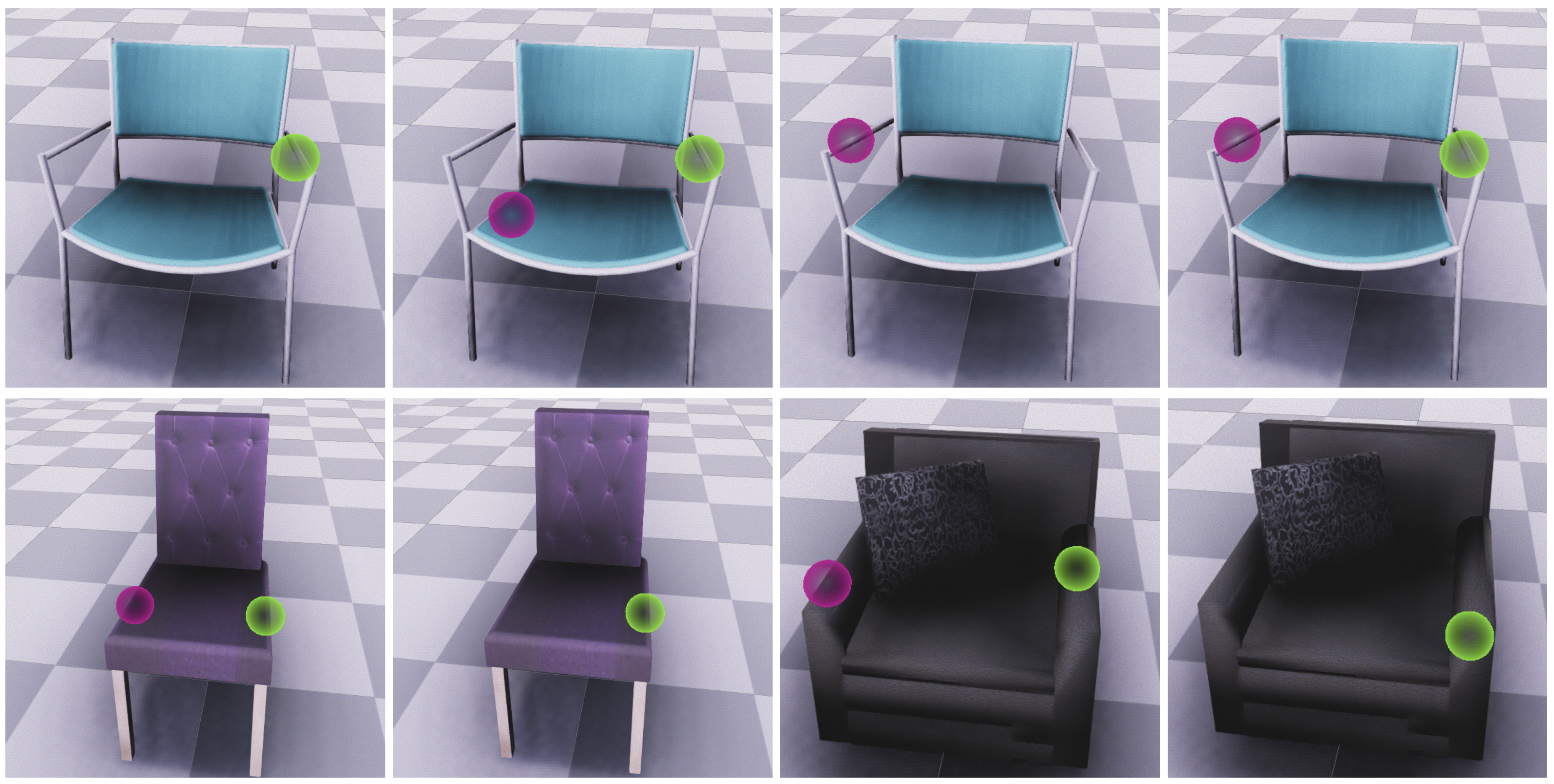}
\caption{\textit{ContactNet} enables sampling of diverse contact positions across various chair shapes. These contacts can be used by our ControlNet and PoseNet to generate fully automatic and diverse motions.}
\label{fig:contact}
\end{center}
\end{figure}
Apart from user-specified contacts, we can additionally generate the contacts on the surface of a given chair using our proposed ContactNet. This allows us to generate fully automatic and diverse motions for sitting. 
To measure the diversity of the generated contacts from ContactNet, we compute the Average Pairwise Distance (APD) among the generated hand contact positions $c_{j}$ with unseen chair shapes. A total number of 200 unseen chairs are chosen, and each 10 contact positions are predicted for both hands. 
\begin{equation}
\begin{split}
APD = \frac{1}{2LN(N-1)}\sum_{k=1}^{2}\sum_{l=1}^{L}\sum_{i=1}^{N}\sum_{j\neq i }^{N}\|X_{i}' - X_{j}'\|_{2}^{2}
\label{eqn:apd2}
\end{split}
\end{equation}
$L = 200$ is the number of objects and $N = 10$ is the number of contacts generated per object. The APD on contact positions is \textbf{11.82} cm which is comparable to the ground truth dataset which has an APD of \textbf{14.07} cm. As shown qualitatively in Figure \ref{fig:contact}, we can generate diverse and plausible contact positions on chairs, which can generalize to unseen shapes.

\section{Conclusion}

% We present \model, a dataset for human chair interactions with an emphasis on hand contacts. It consists of 3 hours of motion capture with 6 subjects interacting with registered 3D chair models, captured in high quality with IMUs and Kinects. In addition to the dataset, we propose the method, \model, the first method for synthesising controllable contact-based human-chair interactions. Given just initial conditions and the contacts on the chair, our model adopts hand motion planning and full-body motion prediction to achieve fine-grained control over the generated animations. It is demonstrated, our method consistently outperforms SoTA method by improving the average contact accuracy by $\sim$55$\%$ to better satisfy contact constraints. Moreover, it can be seen in the supplementary video that our approach generates more natural motion compared to the baseline methods. We further discuss the limitations and future research in this direction in the supplementary material. Our dataset and code will be released for further research in this direction.
We propose \model, the first method for synthesising controllable contact-driven human-chair interactions. Given initial conditions and the contacts on the chair, our model plans the motion of the hands, which drives the full body poses to satisfy contacts.
In addition to the model, we contribute the \model dataset for human chair interactions which includes a wide variety of sitting motions approaching and contacting the chair in different ways. 
It consists of 3 hours of motion capture with 6 subjects interacting with registered 3D chair models, captured in high quality with IMUs and Kinects.
 Experiments demonstrate that our method consistently outperforms the SoTA by improving the average contact accuracy by $\sim$55$\%$ to better satisfy contact constraints. In addition to better control, it can been seen in the supplementary video that our approach generates more natural motion compared to the baseline methods. 
 In the future, we want to extend our dataset to new activities and train a multi-activity contact driven model. In the supplementary, we discuss further future directions in this new problem of fine-grained controlled motion synthesis. 
 %We further discuss the limitations and future research in this direction in supplementary material. 
 Our dataset and code will be released to foster further work in this new research direction.

% Currently, this work only investigates the chair interaction and hand contacts are studied as a test-bed for controllable human-chair interactions. It would be useful to extend this in the future to a wider range of objects, and dealing with non-static objects such as lifting or moving an object, by conditioning on different contact points. And extend the end effectors control to more than just hands but feet as well.

\section*{Acknowledgement}
We would like to thank Xianghui Xie for helping with data processing, and we are very grateful for all the participants who took part in the data capture.

\newpage
\bibliographystyle{splncs04}
\bibliography{0_Main}
% \end{document}
% \begin{document}
% \input{10_supplementary}
\newpage
\section*{APPENDIX}

In this appendix, we provide additional information about the dataset, implementation details, post-processing techniques. We also discuss on the current limitation as future research perspectives.

\section*{1\quad Dataset}

\subsection*{1.1\quad Motion Data}
Table \ref{table:dataset} shows a break down of our dataset in terms of different types of interactions. Our dataset consists of 3 hours of MoCap with over 500 motion sequences.

\begin{table}[h]
    \small
    	\caption{Distribution of the \model~dataset with different types of interaction.} 
	\centering
	\begin{tabular}[b]{c|ccc}
		\hline
		{Interaction Type} & {Minutes} & {$\%$}  \\
        \hline
        $\textrm{Right Hand}$ & 36.3 & 17.3 \\
        $\textrm{Left Hand}$ & 29.4 & 14.0  \\
        $\textrm{Both Hand}$ & 60.5 & 28.9  \\
        $\textrm{No Contact}$ & 36.5 & 17.4  \\
        $\textrm{Free Interaction}$ & 31.9 & 15.2 \\
        $\textrm{Locomotion}$ & 15.1 & 7.2 \\

        % $\textrm{Multi-Action Modes}$ & \xmark & \xmark & \cmark \\
    	\hline
	\end{tabular}
    \label{table:dataset}
\end{table}

\subsection*{1.2\quad Data Processing}

\textbf{SMPL Fitting.} 
We segment the human in captured RGB images by running Detectron V2 \cite{wu2019detectron2} followed by manual correction with \cite{fbrs2020} on the segmentation masks. These masks are then used to segment multi-view depth maps and lift human point clouds from 2D to 3D. We use FrankMocap \cite{rong2021frankmocap} to initialize the SMPL pose from the images and then apply instance specific optimization \cite{alldieck19cvpr} to fit the SMPL model to the segmented human point cloud. For more accurate fitting, we additionally obtain the SMPL shape parameters of each subject from 3D scans using \cite{bhatnagar2020ipnet}.\\
\textbf{Synchronization with the IMUs.} 
The fitted SMPL model provides us with accurate contacts with the scene, however, the fitted motion sequence is prone to occlusion and drastic body movements, as a result, the fitted motion can be jittery at times. On the other hand, the pose captured with the IMUs is smooth over time, but it might not accurately capture the contacts. To this reason, we synchronize the Kinect captured data with the body sensors by incorporating the SMPL fitted poses into the IMU pose sequences. After synchronization, we optimize the joint rotations $\poserotation$ to achieve temporal smoothness via the objective
\begin{equation}
\begin{split}
    L_{\mathrm{temp}}(\poserotation) = \sum_{i=1}^{T-1}\|\boldsymbol{j}_{i+1}^{r} - \poserotation \| ^{2} + \sum_{i=1}^{T-1} \|\poseacc \|
 \end{split}
\label{eq:tmp}
\end{equation}
where $\poseacc$ represents the acceleration of the body joints in frame $i$ approximated by central difference. 

We additionally use the binary contact labels of the toes and the heels detected by the IMU sensors to remove foot-sliding on the motion data. To remove the foot-sliding, we compute the average joint positions over the duration of the contacts grouped by the positive contact labels. This computation is performed for all four foot joints. This forms a sequence of target joint positions of the feet $\feettar\in R^{4\times 3}$. We then optimize the objective function
\begin{equation}
\begin{split}
    L_\textrm{slide}(\poserotation)=\sum_{i=1}^{T}\|\feettar - \boldsymbol{f}_{i}\| ^{2},
 \end{split}
\label{eq:slide}
\end{equation}
where $\boldsymbol{f}_{i}$ represents the foot joint positions at frame $i$. The resulting motion sequence is temporally smooth and has accurate contacts registered with the chair models.\\
\textbf{Object Processing.} 
To obtain object segmentation, we pre-scan objects using a 3D scanner \cite{treedys,agrisoft}. We then use multi-view object keypoints, marked by manual annotators on the images, to fit the pre-scanned chair meshes to the given frame. The segmentation masks are then obtained by projecting fitted object meshes to the images. Since the chairs remain static during the capture, we average over the 6D pose of the fitted chair model during each capture session to obtain the final transformation of the chair.

\section*{2\quad Training Details}
\begin{figure}[h!]
\begin{center}
\includegraphics[width=\linewidth]{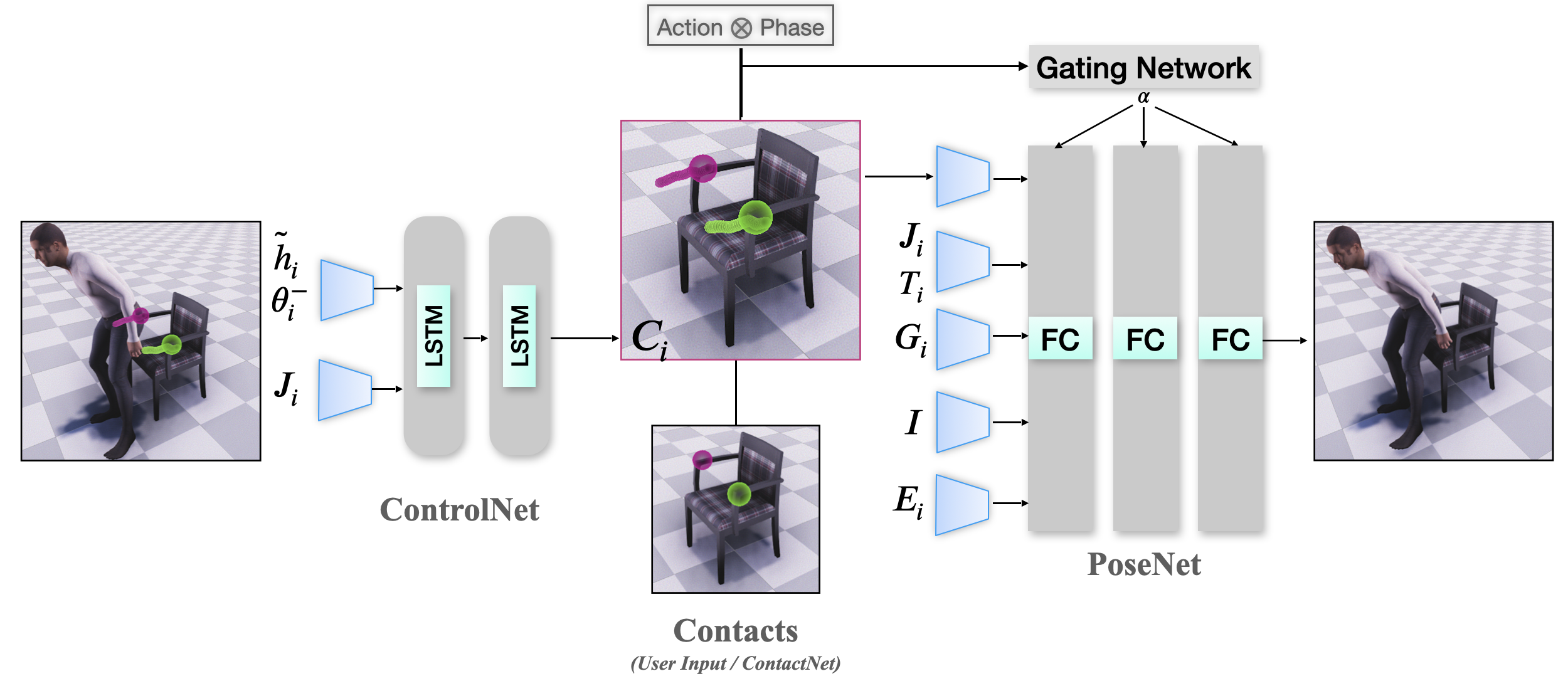}
\caption{Our method that combines the ControlNet and the PoseNet.}
\label{fig:architecture}
\end{center}
\end{figure}
\subsection*{2.1\quad ControlNet}
As shown in Figure \ref{fig:architecture}, the contact network is a two-layer LSTM architecture. Each layer has a hidden dimension of 512. The pose and the control signals (hand trajectories, and the local phases) are each encoded through a two-layer fully connected network with of shape \{128, 128\} before passing through the LSTM. We apply scheduled sampling on hand trajectories for better model performance. For the local phases, we always use the ground truth. Each of our training samples is in a sequence of 60 frames. The ControlNet is trained for 150 epochs with an Adam optimizer. The initial learning rate is 1e-3 and a cosine learning rate scheduler was used to decay the learning rate gradually to 5e-6. The full training of a subject-specific model takes approximately 1 hour on an NVIDIA V100 GPU.

\subsection*{2.2\quad PoseNet}
\begin{table}[h!]
    \small
    	\caption{Details on different encoder networks of the PoseNet.} 
	\centering
	\begin{tabular}[b]{c|ccc}
		\hline
		{Networks} & {Architecture}  \\
        \hline
        \textrm{Encoder} for $\control$ & \{128,128,128\} \\
        \textrm{Encoder} for $\{\pose$, $\traj\}$ & \{512, 512, 512\}   \\
        \textrm{Encoder} for $ \goal$ & \{128,128,128\}  \\
        \textrm{Encoder} for $ \scene$ & \{512, 512, 512\}   \\
        \textrm{Encoder} for $\environment$ & \{256,256,256\} \\

        % $\textrm{Multi-Action Modes}$ & \xmark & \xmark & \cmark \\
    	\hline
	\end{tabular}
    \label{table:posenet}
\end{table}

The PoseNet adopts the mixture-of-expert structure \cite{moe}. It consists of different feature encoders of structures shown in Table \ref{table:posenet}. The gating network and the prediction networks are both three-layer fully-connected networks, with hidden dimensions of 128 and 512 respectively. The number of experts is set to 10. The PoseNet is trained for 150 epochs with an Adam optimizer. The initial learning rate is 1e-4 and a cosine learning rate scheduler was used to decay the learning rate gradually to 5e-6. The full training of a subject-specific model takes approximately 6 hours on an NVIDIA V100 GPU.

\subsection*{2.3\quad ContactNet}

The ContactNet encodes the scene $\scene$ through a three-layer fully connected network of shape \{512, 512, 64\}. The latent vector $\latent$ of the VAE is of size 6. The weight of the Kullback-Leibler divergence $\beta$ is 0.1. We use the Adam optimizer with a learning rate of 1e-3 and train ContactNet for 150 epochs. The full training of a subject-specific model takes approximately 10 minutes on an NVIDIA V100 GPU. 

\section*{3\quad Contact Projection and Trajectory Fitting}
To ensure the ContactNet predicts contacts that land exactly on the surface of the object, we perform a post-processing step, when the distance of the network predicted contact to the surface is less than a set threshold of 10 cm, we simply project the contact onto the nearest point on the chair surface. When the distance is greater than 10 cm, we simply neglect the predicted contact. The ControlNet predicts the future hand trajectories, and it would be possible to fit the predicted pose to the predicted hand position from the hand trajectories at each frame to further improve the satisfaction of the contact constraints. Note, in the evaluation of the main paper we do not apply such fitting technique.

\section*{4\quad Limitations and Future Direction}
We observe the synthesized motion can slightly intersect with the chair. A solution to this problem would be to apply a post-processing step to avoid such collision. In order to generalize to more different chair shapes, it would be useful to investigate better ways of encoding the scene geometry while trying to avoid over-fitting.

Different shaped person can intersect with the same object very differently even when performing the same motion. The \model~dataset captures human interaction with different body shapes. With the dataset, it is possible to study how to build subject-variant motion synthesis model and how to effectively condition on the body shapes. These are challenges in motion synthesis that have not been tackled.

Our work on controllable human-chair interaction. It would be useful to extend the scope of interacted objects, especially considering the cases when the objects are non-static, when performing motions such as lifting a box, or opening a door. Another possible direction would be to further apply contact-based control in these interactions.
\end{document}